\documentclass[conference]{IEEEtran}
\IEEEoverridecommandlockouts
\usepackage{cite}
\usepackage{amsmath,amssymb,amsfonts}
\usepackage{graphicx}
\usepackage{textcomp}
\usepackage{xcolor}
\usepackage{booktabs}
\usepackage{array}
\usepackage{url}
\usepackage{multirow}
\usepackage{enumitem}
\usepackage{tikz}
\usepackage{balance}
\usepackage{placeins}

\newcommand{\methodname}{PragAlign}
\newcommand{\unknown}{\textsc{Unknown}}
\definecolor{directc}{gray}{0.62}
\definecolor{rulec}{gray}{0.38}
\definecolor{pragc}{gray}{0.05}
\definecolor{softgrid}{gray}{0.82}
\definecolor{softbg}{gray}{0.94}
\setlist[itemize]{leftmargin=*,itemsep=1pt,topsep=2pt}
\def\IEEEbibitemsep{0pt plus .5pt}

\begin{document}

\title{PragAlign: Evidence-Sensitive Reply Assistance Across Chinese and Japanese Appropriateness Judgments}

\author{
\IEEEauthorblockN{
Xin Zhong\textsuperscript{\textdagger},
Satori Hachisuka\textsuperscript{*}
}
\IEEEauthorblockA{
The University of Tokyo\\
Emails: zhongxin@g.ecc.u-tokyo.ac.jp
}
}

\maketitle

\begin{abstract}
Reply assistance in multilingual settings requires linguistic competence and culturally situated judgments of appropriateness. We present \methodname, which separates context reading from selective clarification, and evaluate it alongside Direct and Rule. Nine native Chinese speakers judged Chinese materials; three native Japanese speakers judged matched Japanese versions. In Chinese, \methodname\ received significantly better ranks than both baselines. In Japanese, Direct had the lowest mean rank, \methodname\ had the highest top-rank rate, and the omnibus difference was not significant. The groups selected the same top condition in 5 of 10 scenarios, including four shared \methodname\ selections. The results identify shared and language-specific judgment patterns and inform reply assistance designed to support linguistic and cultural understanding.
\end{abstract}


\section{Introduction}
LLMs increasingly draft messages for professors, colleagues, and collaborators, yet fluent wording can still sound evasive, too direct, overly specific, or socially misaligned. Cross-cultural pragmatic failure arises when intended meaning and social interpretation diverge \cite{thomas1983failure}; politeness and rapport also depend on distance, power, imposition, rights, and obligations \cite{brown1987politeness,spencer2008culturally}.

The problem is not only missing cultural knowledge. Culturally grounded data improve norm coverage \cite{li2024culturepark}, but cannot establish whether a recipient was informed, an audience is private, or a deadline is flexible in a particular exchange. Conversely, asking about every missing detail creates an interaction burden. We therefore label reply-relevant fields as observed, inferred, or unknown, and ask at most one question only when the answer could materially change the reply. This connects clarification research \cite{zhang2023clarify} with pragmatic assistance.

We introduce \methodname, a two-module decision layer before a fixed generator. The Context Reader structures the evidence, and the Gap Policy chooses whether to proceed or ask one question. Our broader objective is reply assistance that supports linguistic and cultural understanding. This initial study compares three reply conditions through matched own-language judgments by native Chinese and Japanese speakers.

Native Chinese speakers evaluated Chinese materials, while native Japanese speakers evaluated matched Japanese versions. Thus, the study compares language-matched appropriateness judgments, not the accuracy of composing replies in a non-native language.

We ask: \textbf{RQ1} What ranking patterns emerge for Direct, Rule, and \methodname\ within the Chinese and Japanese language versions? \textbf{RQ2} In which matched scenarios do the two native-speaker groups select the same or different reply condition? \textbf{RQ3} Which judgment criteria explain these shared and divergent preferences?

Our contributions are:
\begin{itemize}[leftmargin=*]
    \item an evidence-sensitive formulation and two-module decision layer for selective clarification before fixed reply generation;
    \item controlled scenarios grounded in pragmatic and social-norm variables; and
    \item a two-language human evaluation that applies the same ranking analysis to both groups and examines convergence and divergence across matched cases.
\end{itemize}

\section{Background and Related Work}
\subsection{Pragmatic Competence and Appropriateness}
Pragmatic competence selects language that fits a social situation. Politeness theory links choice to face, distance, power, and imposition \cite{brown1987politeness}; rapport management adds rights, obligations, and expectations \cite{spencer2008culturally}. These accounts motivate our scenario variables---relationship, channel, audience, responsibility, factual commitment, and urgency---and our refusal to infer a correct reply directly from a cultural label.

\subsection{Culturally Aware LLMs and Controlled Data}
CulturePark uses cross-cultural dialogue data to improve cultural understanding \cite{li2024culturepark}. Controlled resources likewise support analysis of social norms and style: NormDial uses comparable bilingual synthetic dialogues \cite{li2023normdial}, while GYAFC benchmarks formality-sensitive rewriting \cite{rao2018gyafc}. We follow this controlled-data tradition, structuring synthetic scenarios around literature-grounded dimensions and auditing public versus withheld information.

\subsection{Clarification as Interaction}
Clarification research separates when to ask, what to ask, and how to use the answer \cite{zhang2023clarify}. \methodname\ adapts this decomposition, asking only when an unknown field could change content, tone, responsibility, channel, or audience.

\section{Problem Formulation}
Let $x$ denote the public material available before clarification: the user's request, the incoming message, and any explicitly supplied scenario context. The \methodname\ decision layer consists of two modules. The Context Reader maps $x$ to an evidence-tagged frame $z$, and the Gap Policy selects an action $a\in\{\textsc{proceed},\textsc{ask}\}$ and, when needed, one clarification question $q$:
\begin{equation}
z=f_{\theta}(x), \qquad (a,q)=g_{\phi}(x,z),
\end{equation}
where $q=\varnothing$ when $a=\textsc{proceed}$. A fixed generator $G$ then produces the final reply:
\begin{equation}
y=
\begin{cases}
G(x,z), & a=\textsc{proceed},\\
G(x,z,q,v), & a=\textsc{ask},
\end{cases}
\end{equation}
where $v$ is the user's answer to the clarification question. This notation does not introduce a separately trained update module; in implementation, the clarification answer is inserted into the final generation prompt.

Each field in $z$ is labeled \textsc{observed}, \textsc{inferred}, or \textsc{unknown}. A consequential context gap is an unknown whose resolution could change the reply. Preserving \unknown\ prevents unmentioned relationships, audiences, attitudes, or cultural expectations from becoming confident assumptions.

\section{PragAlign Method}
\methodname\ is a pre-generation decision layer, not a replacement generator. Direct sends public input to the generator; Rule adds a generic relationship/channel/tone instruction; \methodname\ supplies a case-specific frame, uncertainty status, and optional clarification answer. The Context Reader emits a compact frame, while the Gap Policy predicts \textsc{proceed} or \textsc{ask} and targets one unknown field. Gap types cover facts, audience, permission, channel, urgency, recipient goal, and reply language.

\begin{figure*}[t]
\centering
\includegraphics[width=.65\textwidth,trim=110 70 70 70,clip]{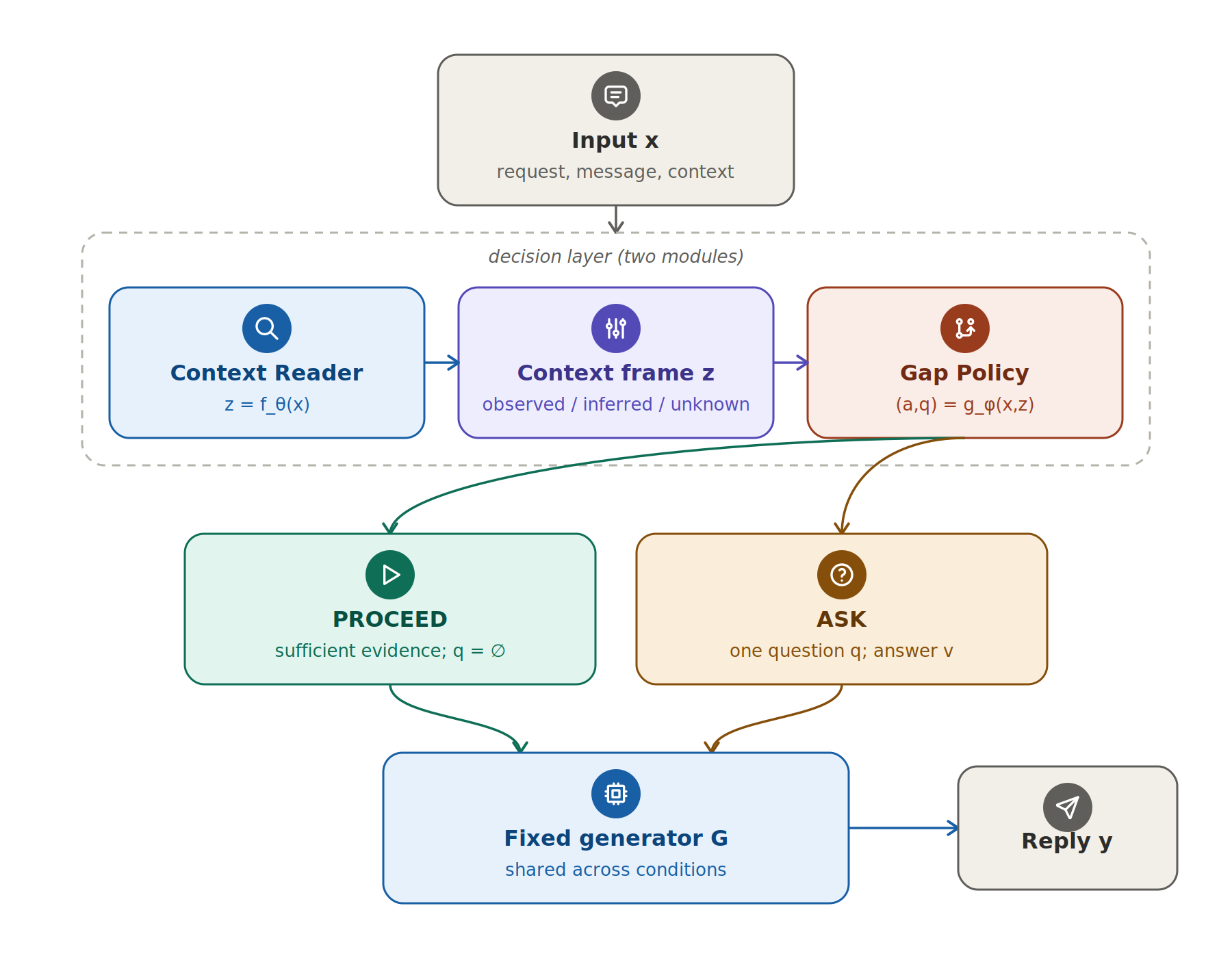}
\caption{PragAlign architecture. The two-module decision layer selects \textsc{proceed} or one targeted clarification before fixed reply generation.}
\label{fig:frame}
\end{figure*}

\section{Data Construction and Model Training}
\subsection{Scenario Construction}
Each controlled pragmatic vignette specifies a request, incoming message, role, relationship, channel, audience, factual constraints, and reply language. These variables follow politeness and rapport theory \cite{brown1987politeness,spencer2008culturally} and controlled social-norm/style resources \cite{li2023normdial,rao2018gyafc}.

Each accepted root yields a matched counterfactual pair: \textsc{proceed} exposes all consequential information, whereas \textsc{ask} withholds one decisive variable while holding other material constant. Supervision thus depends on evidence availability rather than topic or length. Roots are filtered for grounding, public/withheld separation, language consistency, and one answerable gap. These synthetic cases provide auditable supervision, not an estimate of naturally occurring communication.

\subsection{Training Views and Configuration}
View-specific tasks supervise the Context Reader with evidence-tagged frames, and the Gap Policy with the action, the gap type, and one clarification question. Their train/validation/test splits contain 1,520/305/311 and 1,627/343/368 examples, respectively. \textsc{proceed} cases and multiple gap types discourage an always-ask policy.

Both modules initialize from Qwen3-8B \cite{yang2025qwen3} and use LoRA/QLoRA-style parameter-efficient adaptation \cite{hu2022lora,dettmers2023qlora}. The shared final generator is not fine-tuned.

\subsection{Diagnostic Scope}
Held-out counterfactual diagnostics check structured outputs and \textsc{proceed}/\textsc{ask} discrimination. Because synthetic splits may contain lexical cues, the primary evidence comes from human evaluation of final replies.

\section{Human Evaluation}
\subsection{Scenario Set and Procedure}
The evaluation used ten matched scenario specifications: five simple cases with two or three visible reply-relevant constraints, and five complex cases with at least five constraints spanning relationship, audience, responsibility, factual commitment, channel, and urgency. Complexity therefore denotes decision density rather than text length. Each specification was realized naturally in Chinese and Japanese while preserving role, public and withheld information, channel, and allowable facts (Table~\ref{tab:evaluation-design}).

Nine native Chinese speakers evaluated Chinese materials (90 participant--case blocks), and three native Japanese speakers evaluated the matched Japanese versions (30 blocks). Both groups ranked three anonymized, randomized replies from 1 (most appropriate) to 3 (least appropriate) and briefly explained each judgment. The task structure and analysis were identical across versions. The study compares language-matched judgments, not non-native-language reply composition.

\subsection{Statistical Analysis}
For each group, we compare condition-level ranks using Friedman's test \cite{friedman1937ranks} and report mean rank, bootstrap confidence intervals \cite{efron1993bootstrap}, top- and worst-rank rates, and Kendall's $W$ \cite{kendall1939rankings}. Significant omnibus effects are followed by Wilcoxon tests \cite{wilcoxon1945individual} with Holm correction \cite{holm1979simple}. Cross-language comparison uses matched cases rather than pooled blocks.

\subsection{Experimental Controls}
Conditions shared the same scenario facts and fixed generator; only the pre-generation path differed. Labels were hidden and reply order randomized. Both groups were analyzed with the same summary measures and omnibus test.

\begin{table}[t]
\caption{Language-matched evaluation design and Chinese results by scenario complexity. Lower mean rank is better.}
\label{tab:evaluation-design}
\centering
\small
\setlength{\tabcolsep}{5pt}
\renewcommand{\arraystretch}{1.10}
\begin{tabular}{@{}lcc@{}}
\toprule
\textbf{Design item} & \textbf{Chinese version} & \textbf{Japanese version} \\
\midrule
Scenario roots & \multicolumn{2}{c}{10 matched (5 simple, 5 complex)} \\
Materials & Chinese replies & Japanese replies \\
Native speakers & 9 (90 blocks) & 3 (30 blocks) \\
Task & \multicolumn{2}{c}{Rank three replies and explain} \\
\midrule
\multicolumn{3}{@{}l}{\textit{Chinese result by complexity}} \\
\textbf{Condition} & \textbf{Simple} & \textbf{Complex} \\
Direct & 2.27 & 2.36 \\
Rule & 2.16 & 2.00 \\
\textbf{PragAlign} & \textbf{1.58} & \textbf{1.64} \\
PragAlign top-rank rate & 53\% & 49\% \\
\bottomrule
\end{tabular}
\end{table}

\begin{figure}[t]
\centering
\resizebox{\columnwidth}{!}{%
\begin{tikzpicture}[y=.90cm,font=\sffamily\small,line cap=round]
% Panel A: forest plot.
\node[anchor=west,font=\sffamily\bfseries] at (.05,7.30)
  {(a) Mean rank and 95\% CI (lower is better)};
\foreach \x/\lab in {2.80/1.4,5.30/1.8,7.80/2.2}{
  \draw[softgrid,line width=.45pt] (\x,3.52)--(\x,6.82);
  \node[anchor=north] at (\x,3.42){\lab};
}
\node[anchor=west,font=\sffamily\bfseries] at (.10,6.60){Chinese};
\node[anchor=west,font=\sffamily\bfseries] at (.10,5.00){Japanese};
\foreach \y/\lab in {6.22/Direct,5.82/Rule,5.42/PragAlign,4.60/Direct,4.20/Rule,3.80/PragAlign}
  \node[anchor=east] at (2.15,\y){\lab};

\foreach \y/\xl/\xm/\xr/\val/\lw in {
  6.22/7.60/8.55/9.42/2.31/.75,
  5.82/5.91/7.04/8.16/2.08/.75,
  5.42/3.24/4.11/5.05/1.61/1.05,
  4.60/4.24/5.74/7.18/1.87/.75,
  4.20/5.74/7.61/9.49/2.17/.75,
  3.80/4.49/6.36/8.43/1.97/1.05}{
  \draw[line width=\lw pt] (\xl,\y)--(\xr,\y);
  \draw[line width=.55pt] (\xl,\y-.11)--(\xl,\y+.11);
  \draw[line width=.55pt] (\xr,\y-.11)--(\xr,\y+.11);
  \fill (\xm,\y) circle (.085);
  \node[anchor=west] at (9.68,\y){\val};
}
\draw[line width=.45pt] (2.80,3.52)--(9.55,3.52);

% Panel B: top-rank rate (shifted down for separation from panel A).
\node[anchor=west,font=\sffamily\bfseries] at (.05,2.72)
  {(b) Top-rank rate (higher is better)};
\foreach \x/\lab in {3.10/0,5.20/20,7.30/40,9.40/60}{
  \draw[softgrid,line width=.50pt] (\x,-1.12)--(\x,2.30);
  \node[anchor=north] at (\x,-1.22){\lab\%};
}
\node[anchor=west,font=\sffamily\bfseries] at (.10,2.10){Chinese};
\node[anchor=west,font=\sffamily\bfseries] at (.10,.42){Japanese};
\foreach \y/\lab in {1.72/Direct,1.32/Rule,.92/PragAlign,.02/Direct,-.38/Rule,-.78/PragAlign}
  \node[anchor=east] at (2.45,\y){\lab};
\foreach \y/\x/\val/\lw in {1.72/4.78/16/.75,1.32/6.57/33/.75,.92/8.46/51/1.05,
  .02/6.25/30/.75,-.38/6.25/30/.75,-.78/7.30/40/1.05}{
  \draw[line width=\lw pt] (3.10,\y)--(\x,\y);
  \fill (\x,\y) circle (.075);
  \node[anchor=west] at (\x+.18,\y){\val\%};
}
\draw[line width=.45pt] (3.10,-1.12)--(9.40,-1.12);
\end{tikzpicture}}
\caption{Mean rank with bootstrap confidence intervals and top-rank rate by language.}
\label{fig:condition-results}
\end{figure}

\begin{figure}[t]
\centering
\resizebox{\columnwidth}{!}{%
\begin{tikzpicture}[font=\sffamily\small,line cap=round]
\node[anchor=west,font=\sffamily\bfseries] at (.05,3.00)
  {Winner by matched scenario};
\node[font=\sffamily\itshape] at (3.75,2.34){simple cases};
\node[font=\sffamily\itshape] at (7.75,2.34){complex cases};
\foreach \x/\lab in {2.25/1,3.00/2,3.75/3,4.50/4,5.25/5,6.25/6,7.00/7,7.75/8,8.50/9,9.25/10}
  \node[anchor=south] at (\x,1.62){\lab};
\draw[softgrid,line width=.45pt] (5.75,.35)--(5.75,2.55);
\node[anchor=east,font=\sffamily\bfseries] at (1.90,1.25){Chinese};
\node[anchor=east,font=\sffamily\bfseries] at (1.90,.65){Japanese};
\foreach \x/\lab in {2.25/R,3.00/P,3.75/P,4.50/P,5.25/P,6.25/P,7.00/P,7.75/P,8.50/R,9.25/D}
  \node[font=\sffamily\bfseries] at (\x,1.25){\lab};
\foreach \x/\lab in {2.25/R,3.00/P,3.75/D,4.50/R,5.25/P,6.25/P,7.00/P,7.75/R,8.50/D,9.25/P}
  \node at (\x,.65){\lab};
\node[anchor=west,font=\sffamily\bfseries] at (1.90,-.12){Same winner: 5/10};
\node[anchor=west] at (6.00,-.12){Shared PragAlign wins: 4};
\node[anchor=west,font=\sffamily\footnotesize] at (1.90,-.72)
  {D = Direct, R = Rule, P = PragAlign.};
\end{tikzpicture}}
\caption{Scenario-level convergence across the matched Chinese and Japanese versions.}
\label{fig:matched-winners}
\end{figure}

\subsection{Overall Two-Language Pattern}
With equal language weights, \methodname\ has the lowest mean rank (1.79), highest top-rank rate (45.6\%), and lowest worst-rank rate (24.4\%) (Table~\ref{tab:balanced-aggregate}); Direct and Rule have mean ranks of 2.09 and 2.12. The aggregate weights the two languages equally rather than pooling participant--case blocks, which would give the Chinese group three times as much weight. It is a summary statistic, not a pooled inferential test. Fig.~\ref{fig:condition-results} reports the same measures by language.

\subsection{Chinese-Language Judgment Pattern}
In the Chinese evaluation, \methodname\ achieved a mean rank of 1.61 (95\% CI [1.47, 1.76]) and was ranked first in 51\% of blocks. The overall difference was significant ($\chi^2(2)=22.87$, $p<.001$, Kendall's $W=0.13$). Holm-corrected tests favored \methodname\ over Direct ($p<.001$) and Rule ($p=.002$); Rule and Direct did not differ significantly ($p=.076$). \methodname\ also won 70 of 90 paired comparisons against Direct and 55 against Rule.

The advantage persisted in both complexity strata (Table~\ref{tab:evaluation-design}): the three conditions differed for simple ($p=.002$) and complex cases ($p=.003$).

The result was not driven by one rater: \methodname\ had the lowest mean rank for eight of nine Chinese participants and won seven of ten scenarios. Exceptions favored brevity or clearer boundaries, rather than longer replies.

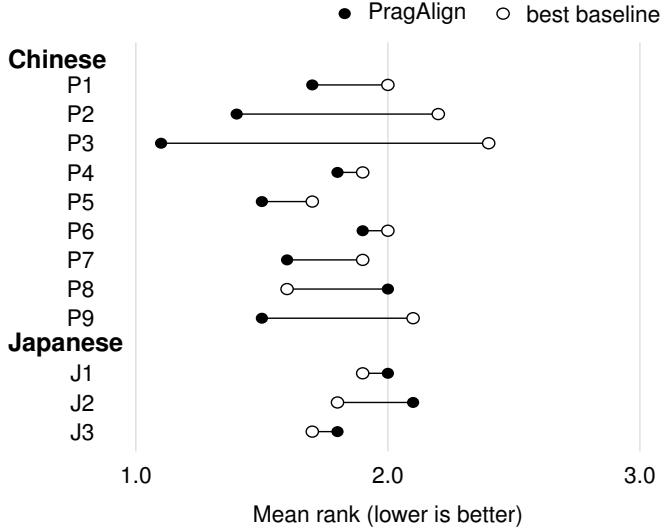
\begin{figure}[t]
\centering
\resizebox{\columnwidth}{!}{%
\begin{tikzpicture}[y=.90cm,font=\sffamily\small,line cap=round]
\fill (4.90,7.05) circle (.085);
\node[anchor=west] at (5.10,7.05){PragAlign};
\draw[line width=.55pt,fill=white] (7.10,7.05) circle (.085);
\node[anchor=west] at (7.30,7.05){best baseline};
% mean rank v maps to x = 2.00 + (v - 1) * 3.5
\foreach \x/\lab in {2.00/1.0,5.50/2.0,9.00/3.0}{
  \draw[softgrid,line width=.40pt] (\x,.30)--(\x,6.60);
  \node[anchor=north] at (\x,.20){\lab};
}
\node[anchor=west,font=\sffamily\bfseries] at (.10,6.35){Chinese};
\foreach \y/\lab in {5.95/P1,5.50/P2,5.05/P3,4.60/P4,4.15/P5,3.70/P6,3.25/P7,2.80/P8,2.35/P9}
  \node[anchor=east] at (1.55,\y){\lab};
\foreach \y/\xp/\xb in {
  5.95/4.45/5.50,
  5.50/3.40/6.20,
  5.05/2.35/6.90,
  4.60/4.80/5.15,
  4.15/3.75/4.45,
  3.70/5.15/5.50,
  3.25/4.10/5.15,
  2.80/5.50/4.10,
  2.35/3.75/5.85}{
  \draw[line width=.55pt] (\xp,\y)--(\xb,\y);
  \fill (\xp,\y) circle (.085);
  \draw[line width=.55pt,fill=white] (\xb,\y) circle (.085);
}
\node[anchor=west,font=\sffamily\bfseries] at (.10,1.95){Japanese};
\foreach \y/\lab in {1.50/J1,1.05/J2,.60/J3}
  \node[anchor=east] at (1.55,\y){\lab};
% J1: PragAlign 2.00, best baseline (Rule) 1.90
% J2: PragAlign 2.10, best baseline (Direct) 1.80
% J3: PragAlign 1.80, best baseline (Direct) 1.70
\foreach \y/\xp/\xb in {
  1.50/5.50/5.15,
  1.05/5.85/4.80,
  .60/4.80/4.45}{
  \draw[line width=.55pt] (\xp,\y)--(\xb,\y);
  \fill (\xp,\y) circle (.085);
  \draw[line width=.55pt,fill=white] (\xb,\y) circle (.085);
}
\node[anchor=north] at (5.50,-.40){Mean rank (lower is better)};
\end{tikzpicture}}
\caption{Participant-level mean ranks in both language groups; participant identifiers are anonymized. \methodname\ had a lower mean rank than the better baseline for eight of nine Chinese participants and none of the three Japanese participants.}
\label{fig:participant-variation}
\end{figure}

\subsection{Japanese-Language and Matched-Case Patterns}
In Japanese, Direct had the lowest mean rank (1.87), while \methodname\ had the highest top-rank rate (40\%). The Friedman test found no overall condition difference ($p=.497$); under the same criterion used for Chinese, pairwise follow-up was therefore not conducted. The discrepancy between mean and top-rank rate shows that \methodname\ was often selected first but not consistently preferred across all positions and scenarios.

The groups selected the same winner in 5 of 10 matched scenarios; four shared winners were \methodname\ (Fig.~\ref{fig:matched-winners}). Agreement occurred in three simple and two complex cases, and shared \methodname\ wins spanned both levels. This is scenario-level convergence rather than inter-rater reliability because each group judged its own-language version. Agreement centered on factual handling and commitments; divergence concerned brevity, boundaries, and explanation length.

\subsection{Qualitative Explanations}
\begin{figure*}[!t]
\centering
\resizebox{\textwidth}{!}{%
\begin{tikzpicture}[x=1.02cm,y=.78cm,font=\sffamily\normalsize,line cap=round]
\fill (6.10,6.15) circle (.105);
\node[anchor=west] at (6.32,6.15){PragAlign ranked 1st};
\draw[line width=.80pt,fill=white] (10.10,6.15) circle (.105);
\node[anchor=west] at (10.32,6.15){PragAlign ranked 3rd};
% Shared category labels
\foreach \y/\lab in {
  4.55/Clarity and concreteness,
  3.65/Politeness and apology,
  2.75/Concise and natural,
  1.85/Too direct or cold,
  .95/Too verbose or specific}{
  \node[anchor=east] at (4.55,\y){\lab};
}
% ---- Panel (a): Chinese. count c maps to x = 5.15 + c * 0.48
\node[anchor=west,font=\sffamily\bfseries] at (5.15,5.35){(a) Chinese};
\foreach \x/\lab in {5.15/0,7.55/5,9.95/10,12.35/15}{
  \draw[softgrid,line width=.50pt] (\x,.48)--(\x,4.90);
  \node[anchor=north] at (\x,.38){\lab};
}
\foreach \y/\xb/\xw in {
  4.55/12.35/6.59,
  3.65/9.95/6.11,
  2.75/7.07/5.15,
  1.85/6.11/6.59,
  .95/5.15/5.63}{
  \draw[line width=.85pt] (\xb,\y)--(\xw,\y);
  \fill (\xb,\y) circle (.115);
  \draw[line width=.85pt,fill=white] (\xw,\y) circle (.115);
}
% Nonzero count labels only; markers on the zero line need no label.
\foreach \x/\y/\lab in {12.35/4.72/15,9.95/3.82/10,7.07/2.92/4,6.11/2.02/2}
  \node[anchor=south,font=\sffamily\small] at (\x,\y){\lab};
\foreach \x/\y/\lab in {6.59/4.38/3,6.11/3.48/2,6.59/1.68/3,5.63/.78/1}
  \node[anchor=north,font=\sffamily\small] at (\x,\y){\lab};
% ---- Panel (b): Japanese. count c maps to x = 13.40 + c * 0.72
% Counts (ranked 1st / ranked 3rd): Clarity 1/2, Politeness 5/0,
% Concise 1/0, Too direct 0/1, Too verbose 0/0.
\node[anchor=west,font=\sffamily\bfseries] at (13.40,5.35){(b) Japanese};
\foreach \x/\lab in {13.40/0,14.84/2,16.28/4,17.72/6}{
  \draw[softgrid,line width=.50pt] (\x,.48)--(\x,4.90);
  \node[anchor=north] at (\x,.38){\lab};
}
\foreach \y/\xb/\xw in {
  4.55/14.12/14.84,
  3.65/17.00/13.40,
  2.75/14.12/13.40,
  1.85/13.40/14.12}{
  \draw[line width=.85pt] (\xb,\y)--(\xw,\y);
  \fill (\xb,\y) circle (.115);
  \draw[line width=.85pt,fill=white] (\xw,\y) circle (.115);
}
% Too verbose row: both counts are zero; markers vertically jittered.
\fill (13.40,1.06) circle (.115);
\draw[line width=.85pt,fill=white] (13.40,.84) circle (.115);
\foreach \x/\y/\lab in {14.12/4.72/1,17.00/3.82/5,14.12/2.92/1}
  \node[anchor=south,font=\sffamily\small] at (\x,\y){\lab};
\foreach \x/\y/\lab in {14.84/4.38/2,14.12/1.68/1}
  \node[anchor=north,font=\sffamily\small] at (\x,\y){\lab};
\node[anchor=north,font=\sffamily\small] at (11.40,-.38)
  {Number of coded mentions};
\end{tikzpicture}}
\caption{Themes coded in explanations when PragAlign ranked first or third, using the same coding scheme for both groups: (a) Chinese, (b) Japanese. A comment may receive multiple codes; absolute counts are not directly comparable across panels because the Chinese group contributed three times as many blocks. Unlabeled markers on the zero line denote zero mentions.}
\label{fig:themes}
\end{figure*}
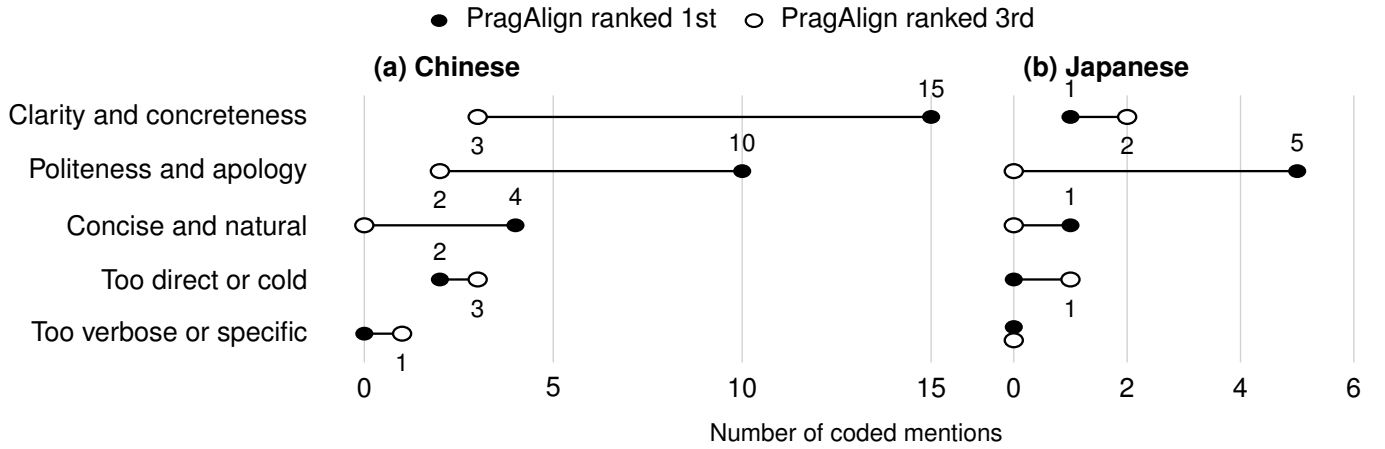

The coded Chinese explanations support the ranking pattern (Fig.~\ref{fig:themes}a): clarity and politeness were concentrated in favorable \methodname\ judgments, whereas unfavorable judgments were more often associated with directness or excess specificity. The Japanese explanations, coded with the same scheme (Fig.~\ref{fig:themes}b), yielded fewer coded mentions overall; favorable judgments centered on politeness, while the few unfavorable mentions concerned clarity and directness.

Because comments could receive multiple codes, the counts profile recurring considerations rather than exclusive categories. In the Chinese data, favorable judgments emphasize actionable commitments and social fit, while unfavorable judgments cluster around length and detail. Thus, the benefit is not reducible to reply length, and over-elaboration remains a failure mode.

\begin{table}[t]
\centering
\caption{Condition-level outcomes. Lower mean/worst ranks are better; higher top-rank rates are better.}
\label{tab:balanced-aggregate}
\footnotesize
\setlength{\tabcolsep}{3.2pt}
\renewcommand{\arraystretch}{1.08}
\begin{tabular}{@{}llccc@{}}
\toprule
\textbf{Eval.} & \textbf{Condition} & \textbf{Mean} & \textbf{Top} & \textbf{Worst} \\
\midrule
Chinese & Direct & 2.31 & 15.6\% & 46.7\% \\
& Rule & 2.08 & 33.3\% & 41.1\% \\
& \textbf{PragAlign} & \textbf{1.61} & \textbf{51.1\%} & \textbf{12.2\%} \\
\midrule
Japanese & Direct & \textbf{1.87} & 30.0\% & \textbf{16.7\%} \\
& Rule & 2.17 & 30.0\% & 46.7\% \\
& PragAlign & 1.97 & \textbf{40.0\%} & 36.7\% \\
\midrule
Equal-lang. & Direct & 2.09 & 22.8\% & 31.7\% \\
aggregate & Rule & 2.12 & 31.7\% & 43.9\% \\
& \textbf{PragAlign} & \textbf{1.79} & \textbf{45.6\%} & \textbf{24.4\%} \\
\bottomrule
\end{tabular}

\vspace{3pt}
\parbox{\columnwidth}{\footnotesize The aggregate weights languages equally. Chinese pairwise wins: 77.8\% vs. Direct, 61.1\% vs. Rule. No Japanese pairwise follow-up (omnibus n.s.).}
\end{table}

\section{Discussion}
\subsection{What the Current Study Can Claim}
The Chinese evaluation showed a significant condition effect, with \methodname\ favored over both baselines. In Japanese, the omnibus effect was not significant, and mean rank and top-rank rate yielded different patterns. The equal-language aggregate is reported only as an equal-weight summary, not a pooled test. Agreement in 5 of 10 matched cases, including four shared \methodname\ selections, identifies common judgments and language-dependent preferences in directness and detail. These findings provide an initial basis for culturally and linguistically informed reply assistance.

\subsection{Implications for Reply Assistance}
Rule is strong because it prompts for relationship, channel, and tone, but a checklist cannot determine whether a field is observed, inferred, unknown, or consequential. \methodname\ makes this epistemic distinction explicit before generation. In the Chinese evaluation, its favorable ranks in both complexity strata suggest that the decision layer can prevent unsupported additions in simple cases and coordinate multiple constraints in complex ones.

The results support a division of labor: the decision layer controls evidence and clarification, while the generator realizes that decision with language-specific brevity, politeness, and explicitness. Rankings measure relative appropriateness; explanations diagnose grounded commitments and over-elaboration.

\subsection{Implications for Additional Experiments}
Evaluating non-native-language reply composition requires a separate shared-language design. Identical materials or bilingual evaluators under controlled uncertainty would better separate language expression from cultural background and support generator comparisons.

\section{Limitations and Future Work}
Synthetic data do not replace natural interactions or human labels. Group sizes differ (nine Chinese and three Japanese participants), reducing between-group precision despite identical analyses. Block-level tests treat participant--case blocks as exchangeable; Fig.~\ref{fig:participant-variation} mitigates but does not remove this concern. The Japanese panels in Figs.~\ref{fig:participant-variation} and~\ref{fig:themes} are reported for presentation parity, but with three participants their per-participant and per-theme counts should be read as descriptive rather than precise. Because each group judged its own-language materials, differences may reflect expression, cultural background, or both. The design does not compare non-native-language reply composition. A preregistered replication should balance groups, use a shared language, independently annotate clarification quality, and test multiple generators.

\FloatBarrier
\section{Conclusion}
This study provides an empirical foundation for culturally and linguistically informed reply assistance. \methodname\ received significantly better ranks than both baselines in Chinese; in Japanese, it had the highest top-rank rate but no significant omnibus effect. Agreement in 5 of 10 matched cases identifies shared and language-specific judgment patterns and motivates a balanced, shared-language follow-up study.

\FloatBarrier
\balance
\bibliographystyle{IEEEtran}
\bibliography{references}

@book{brown1987politeness,
  author = {Brown, Penelope and Levinson, Stephen C.},
  title = {Politeness: Some Universals in Language Usage},
  series = {Studies in Interactional Sociolinguistics},
  number = {4},
  publisher = {Cambridge University Press},
  address = {Cambridge, UK},
  year = {1987},
  doi = {10.1017/CBO9780511813085},
  isbn = {9780521313551}
}

@book{spencer2008culturally,
  editor = {Spencer-Oatey, Helen},
  title = {Culturally Speaking: Culture, Communication and Politeness Theory},
  edition = {2},
  publisher = {Continuum},
  address = {London, UK},
  year = {2008},
  isbn = {9780826493101}
}

@article{thomas1983failure,
  author = {Thomas, Jenny},
  title = {Cross-Cultural Pragmatic Failure},
  journal = {Applied Linguistics},
  volume = {4},
  number = {2},
  pages = {91--112},
  year = {1983},
  doi = {10.1093/applin/4.2.91}
}

@inproceedings{li2024culturepark,
  author = {Li, Cheng and Teney, Damien and Yang, Linyi and Wen, Qingsong and Xie, Xing and Wang, Jindong},
  title = {{CulturePark}: Boosting Cross-Cultural Understanding in Large Language Models},
  booktitle = {Advances in Neural Information Processing Systems},
  volume = {37},
  pages = {65183--65216},
  year = {2024},
  doi = {10.52202/079017-2082}
}

@inproceedings{zhang2023clarify,
  author = {Zhang, Michael JQ and Choi, Eunsol},
  title = {Clarify When Necessary: Resolving Ambiguity Through Interaction with {LM}s},
  booktitle = {Findings of the Association for Computational Linguistics: NAACL 2025},
  pages = {5541--5558},
  address = {Albuquerque, New Mexico},
  publisher = {Association for Computational Linguistics},
  year = {2025},
  month = apr,
  doi = {10.18653/v1/2025.findings-naacl.306},
  isbn = {979-8-89176-195-7}
}

@inproceedings{li2023normdial,
  author = {Li, Oliver and Subramanian, Mallika and Saakyan, Arkadiy and CH-Wang, Sky and Muresan, Smaranda},
  title = {{NormDial}: A Comparable Bilingual Synthetic Dialog Dataset for Modeling Social Norm Adherence and Violation},
  booktitle = {Proceedings of the 2023 Conference on Empirical Methods in Natural Language Processing},
  pages = {15732--15744},
  address = {Singapore},
  publisher = {Association for Computational Linguistics},
  year = {2023},
  month = dec,
  doi = {10.18653/v1/2023.emnlp-main.974}
}

@inproceedings{rao2018gyafc,
  author = {Rao, Sudha and Tetreault, Joel},
  title = {Dear Sir or Madam, May {I} Introduce the {GYAFC} Dataset: Corpus, Benchmarks and Metrics for Formality Style Transfer},
  booktitle = {Proceedings of the 2018 Conference of the North American Chapter of the Association for Computational Linguistics: Human Language Technologies, Volume 1 (Long Papers)},
  pages = {129--140},
  address = {New Orleans, Louisiana},
  publisher = {Association for Computational Linguistics},
  year = {2018},
  month = jun,
  doi = {10.18653/v1/N18-1012}
}

@article{yang2025qwen3,
  author = {Yang, An and others},
  title = {Qwen3 Technical Report},
  journal = {arXiv preprint arXiv:2505.09388},
  year = {2025},
  eprint = {2505.09388},
  archivePrefix = {arXiv},
  primaryClass = {cs.CL}
}

@inproceedings{hu2022lora,
  author = {Hu, Edward J. and Shen, Yelong and Wallis, Phillip and Allen-Zhu, Zeyuan and Li, Yuanzhi and Wang, Shean and Wang, Lu and Chen, Weizhu},
  title = {{LoRA}: Low-Rank Adaptation of Large Language Models},
  booktitle = {International Conference on Learning Representations},
  year = {2022}
}

@inproceedings{dettmers2023qlora,
  author = {Dettmers, Tim and Pagnoni, Artidoro and Holtzman, Ari and Zettlemoyer, Luke},
  title = {{QLoRA}: Efficient Finetuning of Quantized {LLM}s},
  booktitle = {Advances in Neural Information Processing Systems},
  volume = {36},
  pages = {10088--10115},
  year = {2023},
  doi = {10.52202/075280-0441}
}

@article{friedman1937ranks,
  author = {Friedman, Milton},
  title = {The Use of Ranks to Avoid the Assumption of Normality Implicit in the Analysis of Variance},
  journal = {Journal of the American Statistical Association},
  volume = {32},
  number = {200},
  pages = {675--701},
  year = {1937},
  doi = {10.1080/01621459.1937.10503522}
}

@article{wilcoxon1945individual,
  author = {Wilcoxon, Frank},
  title = {Individual Comparisons by Ranking Methods},
  journal = {Biometrics Bulletin},
  volume = {1},
  number = {6},
  pages = {80--83},
  year = {1945},
  doi = {10.2307/3001968}
}

@article{holm1979simple,
  author = {Holm, Sture},
  title = {A Simple Sequentially Rejective Multiple Test Procedure},
  journal = {Scandinavian Journal of Statistics},
  volume = {6},
  number = {2},
  pages = {65--70},
  year = {1979},
}

@book{efron1993bootstrap,
  author = {Efron, Bradley and Tibshirani, Robert J.},
  title = {An Introduction to the Bootstrap},
  address = {New York, NY, USA},
  publisher = {Chapman \& Hall},
  year = {1993}
}

@article{kendall1939rankings,
  author = {Kendall, M. G. and Babington Smith, B.},
  title = {The Problem of $m$ Rankings},
  journal = {The Annals of Mathematical Statistics},
  volume = {10},
  number = {3},
  pages = {275--287},
  year = {1939},
  doi = {10.1214/aoms/1177732186}
}
\end{document}